\documentclass[10pt,twocolumn,letterpaper]{article}

\usepackage{cvpr}
\usepackage{times}
\usepackage{epsfig}
\usepackage{graphicx}
\usepackage{amsmath}
\usepackage{amssymb}
\usepackage{float}
\usepackage{subcaption}
\usepackage{booktabs}
\usepackage{multirow}
\usepackage{pbox}
\usepackage{makecell}
\usepackage{epstopdf}

\usepackage[breaklinks=true,bookmarks=false]{hyperref}

\cvprfinalcopy 

\def\cvprPaperID{****} 

\begin{document}

\title{Learning Efficient Detector with Semi-supervised Adaptive Distillation}

\author{Shitao Tang\qquad Litong Feng\qquad Zhanghui Kuang\qquad  Wenqi Shao\qquad \\ Quanquan Li\qquad Wei Zhang  \qquad Yimin Chen\\Sensetime Research\\ {\tt\small shitaot@gmail.com}}
\maketitle

\begin{abstract}
Knowledge Distillation (KD) has been used in image classification for model compression. However, rare studies apply this technology on single-stage object detectors. Focal loss shows that the accumulated errors of easily-classified samples dominate the overall loss in the training process. This problem is also encountered when applying KD in the detection task. For KD, the teacher-defined hard samples are far more important than any others. We propose ADL to address this issue by adaptively mimicking the teacher's logits, with more attention paid on two types of hard samples: hard-to-learn samples predicted by teacher with low certainty and hard-to-mimic samples with a large gap between the teacher's and the student's prediction. ADL enlarges the distillation loss for hard-to-learn and hard-to-mimic samples and reduces distillation loss for the dominant easy samples, enabling distillation to work on the single-stage detector first time, even if the student and the teacher are identical. Besides, ADL is effective in both the supervised setting and the semi-supervised setting, even when the labeled data and unlabeled data are from different distributions. For distillation on unlabeled data, ADL achieves better performance than existing data distillation which simply utilizes hard targets, making the student detector surpass its teacher. On the COCO database, semi-supervised adaptive distillation (SAD) makes a student detector with a backbone of ResNet-50 surpasses its teacher with a backbone of ResNet-101, while the student has half of the teacher's computation complexity. The code is avaiable at \hyperlink{https://github.com/Tangshitao/Semi-supervised-Adaptive-Distillation}{https://github.com/Tangshitao/Semi-supervised-Adaptive-Distillation}.
\end{abstract}


\section{Introduction}
Boosted by the development of deep convolutional neural network (CNN), the accuracy of object detection has been improved greatly~\cite{liu2016ssd,ren2015faster,lin2017feature}. Along with the requirement of high performance for a detector, low latency is demanded in wide-range practical applications, \eg, mobile apps and autonomous cars. There exists many previous work to speed up CNN, including detection pipeline optimization~\cite{liu2016ssd,ren2015faster,lin2017feature}, architecture design~\cite{howard2017mobilenets,zhang1707shufflenet}, pruning~\cite{han2015learning}, quantization~\cite{zhou2017incremental}, decomposition~\cite{lebedev2014speeding,tai2015convolutional} and knowledge distillation~\cite{hinton2015distilling}. In this paper, we propose semi-supervised adaptive distillation (SAD) scheme to accelerate network in object detecton track. 

Knowledge distillation encourages the student network to converge to a better solution by mimicking the teacher network's feature maps or soften logits. It has achieved great success on image classification~\cite{hinton2015distilling,romero2014fitnets,yim2017gift}. However, when applying it on object detection, due to the "small" capacity of the student network, it is hard to mimic all feature maps or logits well. Knowledge transferring has been applied in the two-stage detector. Chen \etal~\cite{chen2017learning} proposed a weighted cross-entropy loss to underweight matching errors in background regions. Li \etal~\cite{li2017mimicking} mimicked feature maps between the student and the teacher pooled from the same region proposal and discarded those from uninterested regions. Wei \etal~\cite{wei2018quantization} introduced quantization mimic to reduce the search scope of the student network. All the above previous work attempt to design sophistical rules to focus on mimicking informative neurons of the teacher network. In these work, both the teacher and student detectors are two-stage. The application of KD in the single-stage detector has not been explored yet. 

Compared with the two-stage detector, the single-stage detector needs to process much more samples due to the setting of dense anchors. Without the region proposal network (RPN), sample imbalance between easy and hard samples is a special challenge for the single-stage detector. Most of the samples are easy ones during KD for the single-stage detector. However, these easy samples dominate the KD loss. The lack of guidance from hard samples makes KD inefficient for the single-stage detector. There are two types of samples that are important in the distillation process: (1) Hard-to-mimic samples whose gaps between the student's prediction and the teacher's prediction are large; (2) Hard-to-learn samples whose uncertainties defined by teacher's prediction are large. Both the hard-to-mimic and hard-to-learn samples are relevant with the teacher model and should be paid more attention for an effective distillation in the single-stage detector. Previous hard samples defined in focal loss or online hard example mining (OHEM) are selected through comparing predictions with ground truth\cite{lin2017focal,shrivastava2016training}, which are determined only by the detector itself. Due to the difficulties in supervised training and distillation are from different sources, the imbalance treatment between hard and easy samples should be performed for the supervision loss and the distillation loss separately.  With this motivation, an adaptive distillation knowledge loss (ADL) is proposed in this paper, which pays more attention to teacher-defined hard samples and adaptively adjusts the distillation weights between easy-to-mimic/easy-to-learn and hard-to-mimic/hard-to-learn samples in the distillation process. Besides, ADL is also effective in the self distillation setting~\cite{furlanello2018born} just as knowledge distillation. 

Annotating object detection bounding box is extremely time-consuming, which hinders object detection to be used in wide applications. Previous work~\cite{rosenberg2005semi,radosavovic2017data} has demonstrated that unlabeled data can potentially help image classification and object detection. However, in the knowledge distillation scenario, it is an open question how to extract the knowledge of unlabeled data to guide the student network training. The proposed adaptive distillation knowledge also works well in a semi-supervised setting. Provided with potentially unlimited unlabeled data from internet-scale sources, the teacher can present more knowledge to the student via the augmented transferring set. Data distillation~\cite{radosavovic2017data} expresses knowledge of unlabeled data as the annotations produced by the teacher. However, representing knowledge as hard targets of unlabeled may not be an optimal representation. Most of them can be predicted by teacher with very high confidence, so they can also be easily classified by the student. By contrast, soft targets provided by ADL contains balanced easy and hard samples. Thus, soft targets are proposed to be utilized in the semi-supervised distillation for the single-stage detector.

In the real-world application, unlabeled data are far more than labeled data and their distributions are also different, \ie, most unlabeled images do not contain any targeted object. Thus the efficiency of semi-supervised KD will be affected by the large background images. Given these considerations, we raise a practical problem, how to select the unlabeled data which can transfer knowledge more efficiently. In this paper, we show that a trivial filtering mechanism is effective to address this problem.

We select the state-of-the-art single-stage detector RetinaNet~\cite{lin2017focal} to validate the effectiveness of our proposed ADL. Experiments on standard detection data set COCO verify that the proposed ADL can consistently improve the student network's performance, and explore the knowledge of unlabeled data to help the student network to converge to a better solution. Surprisingly, our student detector with a backbone of ResNet-50 even surpasses its teacher detector with a backbone of ResNet-101, even though the student only has half of computation complexity of its teacher. The student detector ResNet-50 achieves an mAP of 36.7 on COCO \textit{test-dev} while the teacher detector ResNet-101 achieves 36.0 when trained only with labeled data.

In the paper, we make the following contributions:
\begin{itemize}
  \item We design an adaptive knowledge distillation loss, which is able to pay more attention to teacher-defined hard samples and adaptively adjust the distillation weights between easy-to-mimic/easy-to-learn samples and hard-to-mimic/hard-to-learn samples for the single-stage object detector.
  \item We develop the proposed adaptive knowledge distillation in a semi-supervised learning setting. The student even surpass its teacher through the semi-supervised KD. 
  \item In order to improve the efficiency of KD in the semi-supervised setting, a data filtering mechanism is proposed to select transferring set from unlabeled data, when unlabeled data and labeled data have different distributions.
\end{itemize}

\section{Related work}
\textbf{Deep network compression and acceleration} Many works are proposed to accelerate the convolution neural network due to the demand for practice applications. Knowledge transferring is one approach that transfers knowledge from the teacher model to the student model. Previous work explores this area by representing knowledge in different forms. FitNet~\cite{romero2014fitnets} makes the student mimic the full feature maps of the teacher. KD~\cite{hinton2015distilling} proposes to supervise the student by soft targets predicted by the teacher. The probability distribution from the teacher model providing extra information than one-hot targets encoding. Our work is closely related to knowledge distillation.

\textbf{Semi-supervised learning and self training} Semi-supervised learning has been studied for years. The goal is to train a model with labeled and unlabeled data. In \cite{rosenberg2005semi}, experiments show object detector can gain extra improvement by Semi-supervised learning. Another work is data distillation\cite{radosavovic2017data}. It first trains a model with labeled data and then uses the model to make predictions on unlabeled data through multi-transform inference and data transformations. Those operations can improve the performance and generate extra knowledge. Different from data distillation, our work focus on knowledge transferring from the strong teacher to the weak student.

\textbf{Object detection} The object detectors include single-stage and two-stage approaches. The two-stage approach consists of two parts, where the first one generates a sparse set of candidate object proposals and then it is fed to a classification and location subnet for further classification and location regression. Single-stage detector directly forward raw pixels through convolution neural network, getting the final classification and location results. One of major problems existing in both types of detectors is class imbalance. In order to address it, Abhinav \etal~\cite{shrivastava2016training} introduces online hard example mining (OHEM) by selecting the top k samples sorted by the loss in one mini-batch. In contrast to two-stage detector where the region proposal network can reduce the candidate location significantly, the single-stage detector suffers severer class imbalance problem. Different from OHEM, focal loss~\cite{lin2017focal} aims to pay more attention to hard examples than easy examples by multiplying a focal term to the common cross entropy loss. Therefore, our distillation loss design follows the spirit of focal loss.

\textbf{Model compression in object detection} Recently, model compression has been studied to facilitate the application of cnn-based object detector in devices with limited computation resources. Chen \etal~\cite{chen2017learning} utilizes soft targets to guide the student model in both region proposal network and region convolution neural network and balance the positive and negative examples by re-weighting the loss of positive and negative samples. Instead of addressing the class imbalance problem directly, Li \etal~\cite{li2017mimicking} proposes to match the feature map after roi-pooling layer where the candidate regions have been significantly reduced. These methods are designed for two-stage detector and cannot be applied to single-stage detector directly. In contrast, our insightful designed loss is another way to address the class imbalance problems.

\section{Semi-supervised Adaptive Distillation}
\begin{figure*}

    \includegraphics[width=\textwidth]{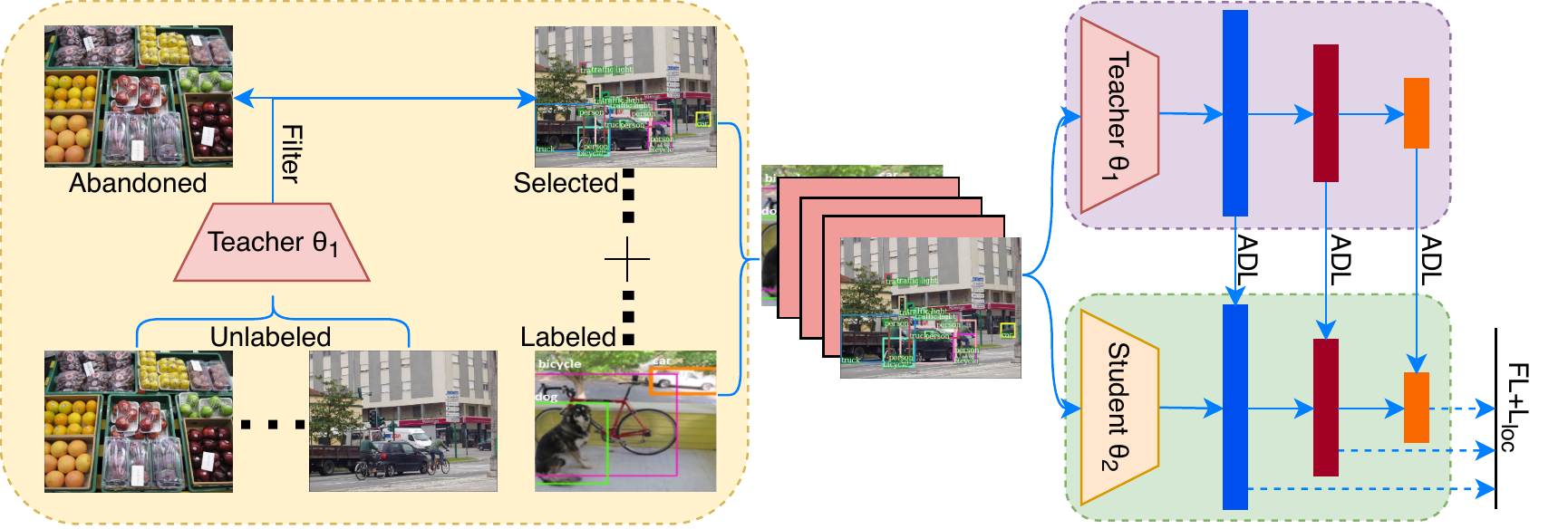}
  \caption{Semi-supervised adaptive distillation (SAD) schematic. To begin with, the teacher selects and annotates samples with at least one annotation. Then, combine those selected samples with labeled ones. At last, the student is trained using \(ADL+FL+L_{loc}\) guided by the teacher. }\label{scheme}
\end{figure*}

The section introduces semi-supervised adaptive distillation (SAD), as shown in Figure \ref{scheme}.
\subsection{Adaptive Distillation}
In this section, we discuss the design of distillation loss for the single-stage detector. Compared with the two-stage detector, the distinguishing feature of the single-stage detector is dense sampling of possible object locations. In a single-stage detector, dense anchors are set on multiple feature maps in the backbone network. Hence distillation needs to be performed on a large number of output logits between teacher and student. In RetinaNet, a typical number of anchors is \(\sim\)100K and most of them correspond to easy-to-mimic or easy-to-learn samples. Though an easy sample contributes little to the distillation loss, the sum of losses from those easy samples will dominate the distillation loss during training. Thus the hard-to-mimic/hard-to-learn samples worth mining are not to be learned well and this restricts the capacity of KD on a single-stage detector yet.

Without loss of generality, we study the case of cross entropy for binary classification. The original focal loss is defined as
\begin{equation}
FL(p_t)=-(1-p_t)^\gamma log(p_t) \label{eq:1}
\end{equation}
\begin{equation}
     p_t=\left\{
                \begin{array}{ll}
                  p, & \text{if y=1}\\
                  1-p, & \text{otherwise}\\
                \end{array}
              \right.
\end{equation}
 In the above, \(y \in \{\pm 1\}\) specified the ground-truth class and \(p \in [0,1]\) is the model's estimated probability for the class with label \(y=1\). 
 
 For the following, We represent \(q\) as the soft probability value predicted by the teacher and \(p\) as the one predicted by the student. Knowledge distillation is inspired by Kullback–Leibler divergence, which  measures the similarity between two distribution, defined as:
 \begin{equation}
     KL(T||S)=q\log(\frac{q}{p})+(1-q)\log(\frac{1-q}{1-p})\label{eq:kl}
 \end{equation}
 The student model tries to mimic the soft class probability distribution predicted by the teacher model. We abbreviate \(KL(S||T)\) as \(KL\) for the following part of the paper. 
 
 \subsubsection{Focal Distillation Loss}
 The common way of adopting focal loss to knowledge distillation is to multiply KL by a focal term. If the focal term utilized by the classification loss (hard targets) is shared by the \(KL\) (soft targets), the joint loss of classification loss and \(KL\) can be defined as
 \begin{equation}
     L=FT(p)(-\log (p_t)+KL)\label{eq:4}
 \end{equation}
  where \(FT(p)\) is the focal term. As shown in Equation (\ref{eq:1}), the focal term is
 \begin{equation}
     FT(p)=(1-p_t)^\gamma
 \end{equation}
 Thus, the focal distillation loss is:
 \begin{equation}
     FDL=(1-p_t)^\gamma KL\label{eq:5}
 \end{equation}
 \(FDL\) is a simple modification from \(FL\) and \(KL\). We consider it as a baseline in our experiments.
 
 \subsubsection{Adaptive Distillation Loss}
 However, as experiments will show, \(FDL\) is dominated by the focal term \(FL\), so \(KL\) contributes little to the overall loss. In order to address the problem, we propose the following loss. We assume that KD on a single-stage detector should focus on measuring the distance of probability distribution between the student and the teacher. Based on this motivation, a modulating factor between 0 to 1 should be used to learn the feature adaptively. Inspired by KL-divergence, We come up with the following term to suit the purpose as:
  \begin{equation}
 DW=(1-e^{-KL})^\gamma
 \end{equation}
 where \(KL\) is defined in Equation (\ref{eq:kl}). DW is abbreviated for distillation weight. Just as the focal loss, the hyperparameter \(\gamma\) controls the rate at which easy examples are down-weighted. The term \((1-e^{-KL})\) controls the weight of each sample. However, \(DW\) only adjusts the weights between students and teachers during the training process. Given that the hard-to-learn samples are extremely important for distillation, we propose \(ADW\) to adjust the percentage of overall weights of hard-to-learn samples (PHLS), defined as:
 \begin{equation}
     ADW=(1-e^{-(KL+\beta T(q))})^\gamma
 \end{equation}
 \begin{equation}
    T(q)=-(q\log(q)+(1-q)\log(1-q))
\end{equation} 
T(q), the entropy of the teacher, reaches maximum when q is \(0.5\) and minimum when q approaches 0 or 1. The teacher probability q reflects the uncertainty of classifying it. When q approaches to \(0.5\), the corresponding sample is treated as a hard-to-learn sample. And a sample with a high KL is treated as a hard-to-mimic sample. Intuitively, PHLS increases when \(\beta\) becomes larger. Thus, \(KL\) controls the weights of hard-to-mimic samples which are adjustable in the training process while \(T(q)\) controls the weights of hard-to-learn samples initially defined by the teacher. The combination of them can adaptively adjust the distillation weights. At last, the adaptive distillation loss is:
 \begin{equation}
     ADL=ADW\cdot KL\label{eq:ADL}
 \end{equation}
 In addition, we will show in the experiment that the above \(ADL\) is effective in the self distillation setting where the teacher and the student are identical.
 
 The original focal loss in one image is the sum of the focal losses over all anchors, normalized by the number of anchors assigned to ground-truth boxes. The proposed adaptive distillation loss for soft targets adopts the same setting of the sum except the normalizer. We found that the training is unstable when the \(ADL\) is normalized by the the normalizer of the original focal loss, because mimicking the soft targets of negative samples predicted by the teacher model also contributes to \(ADL\). In addition, the number of positive samples is unknown for the unlabeled data in the semi-supervised setting. To make the KD training more stable and robust, we define the normalizer as:
 \begin{equation}
     N=\sum_{i}^{n} q_i^\theta\label{eq:7}
 \end{equation}
 The mark q correspond to soft targets of positive samples predicted by the teacher. To be more specific, \(N\) is the sum of probability of positive samples powered by \(\theta\) over all anchors. The modulating factor \(\theta\) reduces the weight contribution from negative samples. In sum, the model is trained with the sum of \(ADL\) over all anchors divided by the \(N\). \(\theta\) is set to 1.8 empirically in all experiments. 
 
\subsubsection{Loss for Distilling Student Model} \label{sec_loss}
 For the student model, we optimize the following function.
 \begin{equation}
     L=FL+ADL+L_{loc}
 \end{equation}
 \(FL\) is the original focal loss and \(L_{loc}\) is the bounding box loss. \(ADL\) is the proposed adaptive distillation loss.

\subsection{Semi-supervised Adaptive Distillation Scheme}\label{unlabel_sec}
 In general, only improving the transferring efficiency by \(ADL\) is hard to bridge the gap between the teacher and the student. Therefore, we try to further improve distillation performance by mining the data efficiently. Labelling samples by human is expensive, especially for the object detection task. The collection of web-scale unlabeled samples is economical. Hence exploiting unlabeled data for KD is investigated here. Previous work~\cite{radosavovic2017data} introduces semi-supervised data distillation, in which the learner exploits all available labeled data plus internet-scale sources of unlabeled data. It reveals a strong connection between the improvement of the student model and the amount of unlabeled data used. Essentially, semi-supervised learning improves distillation performance by augmenting the transferring set. In the semi-supervised setting, the teacher is trained on a labeled dataset \(S_L\) and the student is trained using both \(S_L\) and an unlabeled dataset \(S_U\). The \(S_L\) and \(S_U\) compose the  transferring dataset \(S_T\). The student is guided by soft targets from the teacher on \(S_T\) and ground truth on \(S_L\) simultaneously. KD aims to improve transferring efficiency given that \(S_L=S_T\) in the common setting. We suppose that the expansion of the transferring set can improve KD further. 

 \textbf{Generating labels on unlabeled data} Data distillation proposes to use final output from the teacher model as the distillation model. These labels can be generated from the soft targets according a mapping function. In addition, they are the easiest given that the soft target probability of most of them is greater than 0.95. However, the hard samples dropped by non maximum suppression are of great importance in knowledge transferring. Thus, we propose to use the combination of hard targets and soft targets. The steps are as follows.
 
 (1) Train the teacher model with labeled data; 
 
 (2) Generate hard targets for unlabeled data using the teacher model; 
 
 (3) Train the student model with both labeled and unlabeled data using both soft targets and hard targets.

 \textbf{Unlabeled data selection} For real-world applications, the unlabeled data and the labeled data are often not in the same distribution. Different types of data are collected for different purposes, \ie, ImageNet~\cite{deng2009imagenet} for classification and COCO~\cite{lin2014microsoft} for detection. We can simulate a real-world scenario, for a task \(t_i\), only one data-set \(d_j\) contains the label for it. All the rest data-sets contain unlabeled data for the task \(t_i\) and the amount of those unlabeled data is much larger than that of the labeled data. Therefore, it is inefficient to utilize all the available data to distill the student. For unlabeled data collected from different sources, most of them do not contain any annotations, which can be easily classified as negative by a well-trained model. On the contrary, the image containing at least one positive samples are generally harder to detect. Given these considerations, we propose to select those images which have at least one annotation produced by the teacher to distill the student for transferring knowledge more efficiently.

 \section{Experiment}
 We evaluate our methods on the detection task of the COCO benchmark~\cite{lin2014microsoft}.We report all studies by evaluating on the \textit{mini-val} (5k images) or \textit{test-dev} (41k images) split with the standard metrics of average precision following the COCO definitions, including \(AP, AP_{50},AP_{75}\).

\subsection{Data Splits}
 In the 2017 version of COCO, there exists 115k labeled images and 120k unlabeled images. We referred them as co-115 and un-120 respectively. We train the teacher model on co-115 and the student model on the union of co-115 and un-120.

\textbf{Optimization} We evaluate our method using RetinaNet, one of the state-of-the-art single stage detectors. All the hyper-parameters are the same as \cite{lin2017focal}. We utilize the implementation from detectron~\cite{Detectron2018}. All the models are trained with synchronized SGD over 8 GPUs with a total of 16 images per minibatch (2 images per GPU). The initial learning rate is set as 0.01, weight decay as 0.0001, and momentum as 0.9. For training models only using co-115, we set the iteration size to 90000. For training models on both co-115 and un-120, a iteration size of 270000 is used. The learning rate is divided by 10 at 70\% and 90\% of the total number of iterations. Further increasing the number of iterations will not improve the performance. 

\textbf{Loss} We use the loss introduced above for knowledge distillation. \(\gamma\) is set the same for soft target loss and hard target loss. The classification loss is applied to all ~100k anchors in each sampled image. 

\subsection{Student-teacher Pairs}
We validate our methods in different student-teacher pairs. 

\textbf{Distillation over scales}
We first investigate the performance improvement through distillation when the input size is reduced. For the implementation of KD, the teacher and the student should have the same number of output logits, though input sizes are different. Therefore, We simply add a deconvolutional layer on top of the final feature map of the student model to match the size of the teacher's final feature map. In experiments, the input size of the student model is 400*677 and the input size of the teacher model is 800*1333. Both the teacher and the student utilize ResNet-50 as the backbone.

\textbf{Distillation over small models}
In addition to distillation over different input sizes, we also examine distillation over detectors with different capacities, in which a strong teacher model distills a weak student model. Experiments with several pairs of teacher and student are conducted in this study.

\subsection{Adaptive Distillation Study}

In this section, we compare different methods. We use ResNext-50 as the teacher and ResNet-50-half as the student if not specified. The scale is 800*1333.

\textbf{Feature map mimic} First we evaluate the method of naive logits mimic using L2 loss. The entire feature map regression is implemented through the mimic mentioned in \cite{li2017mimicking}. Results of logits mimic and entire feature map mimic are shown in Table \ref{table:6}. The mimicked models do not obtain any improvement compared to the baseline. The small network is difficult to learn from the teacher through this method.

\textbf{Focal Knowledge Distillation} We try another loss function , which adopts the same focal term between hard targets and soft targets. Surprisingly, the performance of student model drops from 34.3 to 33.9. We attribute the performance decrease to the reason that the supervisions of the ground truth in the focal term is so strong that it neglects the effect of the soft targets. The gradient is 0 when p is equal to the ground truth. In other words, \(FDL\) is not minimum when p is equal to q.

\textbf{Adaptive distillation loss} Results using proposed \(ADL\) with varying \(\beta\) are shown in Table. \ref{varybeta}. When \(\beta\) is 0,which is equivalent to \(DW\), \(ADL\) does not work, since PHLS is very small. The performance improves as \(\beta\) becomes larger. With \(\beta=1.5\), ADL yields nearly 2 AP improvement of the student. Compared with \(FDL\), our proposed \(ADL\) has the property that the loss is minimum when the output p produced by the student is equal to q produced by the teacher. Without specific noticing, the aberration of \(ADL\) means the loss defined in Equation (\ref{eq:ADL}) and we use \(\beta=1.5\) for all the following experiments. 

\textbf{ADL under different student-teacher pairs} In Table \ref{table:FDL2}, we show distillation results using co-115 over different student-teacher pairs, with \(ADL\) used. The performance of student models improves significantly with distillation, despite architectural differences between teacher and student. In general, the weak student model achieves over 1\% improvement in mAP and 2\% in AP50. As the results show, simply adding a deconvolution layer on the top of classification and location subnets will harm the performance compared to the results (30.5 mAP) reported in \cite{lin2017focal} , but it can outperform the normal 400*677 model after distillation.

\begin{table}
    \centering
    \begin{tabular}{ c|c|c|c }
     \hline
     \(\beta\) & AP & AP50&AP75 \\
     \hline
     baseline&28.8&45.8&30.6\\
     0& 28.9&45.9&30.6 \\
     0.5& 29.4&46.3&31.2 \\
     1.0&30.5&48.5&32.7 \\
     1.5&30.7&48.8&32.7\\
     \hline
    \end{tabular}
    \caption{Varying \(\beta\) of ADL. The performance increases as \(\beta\) (PHLS) becomes larger. }\label{varybeta}.
\end{table}

\begin{table}
    \centering
    \begin{tabular}{ c|c|c|c }
     \hline
     Method & AP & AP50&AP75 \\
     \hline
     baseline&28.8&45.8&30.6 \\
     Feature map mimic&28.8&45.8&30.6 \\
     FDL&28.5&45.5&30.2\\
     ADL&\textbf{30.7}&\textbf{48.8}&\textbf{32.7} \\
     \hline
    \end{tabular}
    \caption{Results for different distillation methods. Feature map mimic is to minimize L2 loss between the student and the teacher. \(FDL\) is introduced in Equation (\ref{eq:5}). ADL is introduced in Equation (\ref{eq:ADL}).}\label{table:6}.
\end{table}

\begin{table}
    \centering
    \begin{tabular}{ c|c|c|c|c }
     \hline
     Model &scales& AP & AP50&AP75 \\
     \hline
     T (ResNet-50)&800& 35.4&54.6&37.9\\
     S (ResNet-50 up) &400& 29.8&48.8&30.9 \\ 
     AD &400&\textbf{31.2}&\textbf{50.9}&\textbf{32.5} \\
     \hline
     T (ResNet-50)&800& 35.4&54.6&37.9\\
     S (ResNet-50 half) &800& 28.8&45.8&30.6 \\
     AD&800&\textbf{30.7}&\textbf{48.8}&\textbf{32.7}\\
     \hline
     T (ResNext-101)&600&37.9&57.2&40.6\\
     S (ResNet-50) &600& 34.3&53.2&36.9 \\
     AD&600&\textbf{35.2}&\textbf{54.1}&\textbf{37.7}\\
     \hline
    \end{tabular}
    \caption{Distillation with co-115k using \(FDL\) over different student-teacher pairs. S stands for student and T stands for teacher. The notation is the same in the following table.}
    \label{table:FDL2}
\end{table}

\subsection{Semi-supervised Adaptive Distillation Study}

\begin{table*}[]
\centering
\small
\begin{tabular}{l|l|llll|lll}
\hline
Student (scale)                     & Teacher (scale)                  & co-115 GT & co-115 ST & un-120 HT & un-120 ST & AP  & AP50 & \multicolumn{1}{l}{AP75} \\ \hline
\multirow{4}{*}{ResNet-50 up (400)} & \multirow{4}{*}{ResNet-50 (800)} & \checkmark          &            &            &            & 28.8 & 45.8 & 30.6                      \\
                                    &                                  & \checkmark          &            & \checkmark          &            & 32.1&51.6&33.9                      \\
                                    &                                  & \checkmark          & \checkmark          &            & \checkmark          & 32.3&51.3&34.1                      \\
                                    &                                  & \checkmark          & \checkmark          & \checkmark          & \checkmark          &\textbf{33.2}  &\textbf{53.2}  &\textbf{35.1}                       \\ \hline
\multirow{4}{*}{ResNet-50 half  (800)} & \multirow{4}{*}{ResNet-50 (800)} & \checkmark          &            &            &            & 28.8&45.8&30.6                      \\
                                    &                                  & \checkmark          &            & \checkmark          &            & 32.1&50.6&34.2                      \\
                                    &                                  & \checkmark          & \checkmark          &            & \checkmark          &32.3&50.3&34.6                      \\
                                    &                                  & \checkmark          & \checkmark          & \checkmark          & \checkmark          &\textbf{33.1} &\textbf{52.1} &\textbf{35.2}                       \\ \hline
\multirow{4}{*}{ResNet-50  (600)} & \multirow{4}{*}{ResNext-101 (600)} & \checkmark          &            &            &            & 34.3&53.2&36.9                      \\
                                    &                                  & \checkmark          &            & \checkmark          &            & 35.6&54.7&37.9                      \\
                                    &                                  & \checkmark          & \checkmark          &            & \checkmark          &35.9&54.9&38.5                      \\
                                    &                                  & \checkmark          & \checkmark          & \checkmark          & \checkmark          &\textbf{36.6} &\textbf{55.8} &\textbf{38.9}                       \\ \hline
\end{tabular}
\caption{Distillation results using un-120 under different settings. Ground truths are abbreviated as GT. Soft targets are abbreviated as ST. Hard targets are abbreviated as HT, representing hard targets produced by the teacher. The notation co-115 is the 115k COCO training set with annotation while co-120 is the 120k COCO unlabeled set.}\label{table:un120}
\end{table*}

We conduct different experiments to study distillation with the un-120. The results are summarized in Table \ref{table:un120}. 

\textbf{Experiment setting} The notation co-115 ST or un-120 ST represents soft targets produced by the teacher. Ground truths are abbreviated as GT while hard targets are abbreviated as HT. Hard targets are predicted by the teacher using the method introduced in \cite{radosavovic2017data}.

\textbf{Effect of un-120} First, we investigate the method introduced in \cite{radosavovic2017data}. Following the protocol, we generate annotations for un-120k by selecting a threshold that makes 'the average number of annotated instances per unlabeled image' roughly equal to 'the average number of instances per labeled image'. Compared with the one only using co-115 GT, the use of un-120 yields significant improvement. 


\textbf{Effect of soft targets and hard targets in un-120} We use both soft targets and hard targets of un-120 in the training stage. It is noted that the soft targets and hard targets are all from the teacher model. Compared with the distillation model using only hard targets of un-120k, the combination of them yields over 1 AP improvement in all student-teacher pairs, which shows the effectiveness of hard samples contained in soft targets in knowledge transferring.

\subsection{Surpass the Teacher}
In this section, we show that the student can surpass the teacher with our ADL under the semi-supervised setting. We discuss this investigation in two categories: the student and the teacher are identical or the teacher has better performance than the student. The former is called as self distillation.

\textbf{Self distillation} In the experiments of self distillation, the teacher and the student are parameterized identically. We conduct experiments with different input scales (400, 500, 600) and different backbone models (ResNet-50, ResNet-101). In Table \ref{table:7}, GT is short for ground truth and ST is short for soft targets. As shown in Table \ref{table:7}, our self distillation method achieves around an improvement of 0.5 on AP compared to the teacher model trained with \(S_T\). The improvement by self distillation is consistent over different input scales and backbone models. We also tested to perform self distillation using both co-115 and un-120, but no improvement is gained by the addition of un-120, which indicates that co-115 is sufficient to transfer knowledge from teacher model to student model with the same capacity.

\begin{table}

    \centering
    \begin{tabular}{ c|c|c|c|c|c }
     \hline
      Model&Scales&Targets& AP & AP50&AP75 \\ 
     \hline
     \multirow{2}{*}{ResNet-50}&\multirow{2}{*}{400}&GT & 30.5&47.5&32.7 \\
     &&ST+GT& \textbf{31.0}&\textbf{48.1}&\textbf{32.9}\\
     \hline
     \multirow{2}{*}{ResNet-50}&\multirow{2}{*}{500}&GT & 32.5&50.9&34.8 \\ 
     &&ST+GT&\textbf{33.3}&\textbf{51.4}&\textbf{35.6}\\
     \hline
     \multirow{2}{*}{ResNet-50}&\multirow{2}{*}{600}&GT & 34.3&53.2&36.9 \\
     &&ST+GT& \textbf{34.7}&\textbf{53.4}&\textbf{37.0}\\
     \hline
     \multirow{2}{*}{ResNet-101}&\multirow{2}{*}{600}&GT&36.0&54.8&38.7\\
     &&ST+GT&\textbf{36.3}&\textbf{55.2}&\textbf{38.9}\\
     \hline
    \end{tabular}
    \caption{Self distillation results. GT represents ground truth and ST represents soft targets. The students is trained with GT and ST using proposed ADL. As the results show, improvement is obtained under different scales and networks even if the student and the teacher are identical.}\label{table:7}
\end{table}

\textbf{The teacher is better than the student} In this experiment, it is shown that when the teacher is better than the student, the student can surpass its teacher by augmenting the transferring set \(S_T\). We conducted experiments with two student-teacher pairs: (ResNet-50, ResNet-101) pair and (ResNet-101, ResNext-101) pair. The teacher is trained on co-115 and the student is trained on the union of co-115 and un-120. The performance of the teachers are 1.7 and 2.2 higher than that of the students respectively, but the student can still beat the teacher by some margin. In Table \ref{table:un120}, the ResNet-50 (600) student detector can reach to a mAP of 36.6 guided by the ResNext-101 teacher, only slightly higher than the one guided by ResNet-101 in this experiment. We suppose the limitation for (ResNet-50, ResNext-101) pair is caused by the limited amount of data in the transferring set. 

\begin{table}
\small
\centering
    \begin{tabular}{ c|c|c|c|c }
     \hline
     Model&Scale&AP & AP50&AP75 \\
     \hline
     T (ResNet-101)&600&36.0&54.8&38.7\\
     S (ResNet-50)&600&34.3&53.2&36.9 \\
     AD&600&\textbf{36.3}&\textbf{55.2}&\textbf{38.9} \\
     \hline
     T (ResNext-101)&500&36.6&55.5&39.3\\
     S (ResNet-101)&500&34.4&52.7&36.9\\
     AD&500&\textbf{36.8}&\textbf{55.7}&\textbf{39.4} \\
     \hline
    \end{tabular}
    \caption{Results on the union of co-115 and un-120 using the proposed adaptive distillation loss. We show that the student can surpass its teacher.}
\end{table}

\begin{table}
    \centering
    \begin{tabular}{c|c|c}
    \hline
         &AP&time  \\
    \hline
    YOLOv2~\cite{redmon2017yolo9000}&21.6&25\\
    SSD321~\cite{liu2016ssd}&28.0&61\\
    DSSD321~\cite{fu2017dssd}&28.0&85\\
    R-FCN~\cite{dai2016r}&29.9&85\\
    SSD513~\cite{liu2016ssd}&31.2&125\\
    DSSD513~\cite{fu2017dssd}&33.2&156\\
    FPN-FRCN~\cite{lin2017feature}&36.2&172\\
    \hline
    RetinaNet-50-400&30.5&69\\
    RetinaNet-50-600&34.3&90\\
    RetinaNet-50-800&35.7&123\\
    RetinaNet-101-500&34.4&90\\
    RetinaNet-101-800&37.8&190\\
    \hline
    SAD (ours)&\textbf{36.7}&90\\
    SAD (ours)&\textbf{36.9}&90\\
    \hline
    \end{tabular}
    \caption{Speed (ms) versus accuracy (AP) on COCO \textit{test-dev} , which has no public labels and requires evaluation on servers. Our detector has achieves an AP of 36.9, running at 90 ms per image. The distilled detector is more accurate and faster than RetinaNet-50-800.}
    \label{tab:comparisionwithdetectors}
\end{table}

\textbf{Comparison with other detectors} We compare our distilled detector with different detectors. As shown in Table \ref{tab:comparisionwithdetectors}, our detector is significantly better than any other detector except FPN-FRCN and RetinaNet-101-800. With a comparable mAP with the above two detectors, our distilled detectors run 2\(\times\) faster. 


\subsection{Distillation with Dissimilar-distribution Data}
In this section, we evaluate the proposed unlabeled data selection method. Unlabeled data are from the ImageNet dataset other than the COCO dataset. Thus unlabeled data and labeled data have different distributions. 

\textbf{Configuration} First, 110k images are randomly selected from ImageNet images that contain at least one annotation predicted by the teacher. Another 110k images are randomly selected from ImageNet images that do not contain any annotation predicted by the teacher. The threshold of the teacher's prediction is kept the same as the one utilized for un-120. We refer the former as ImageNet\(_p\)-110 and the later as ImageNet\(_n\)-110, in which the subscript \(p\) means positive responses from the teacher and \(n\) means negative responses. We choose the (ResNet-50, ResNext-101) pair in this experiment.

\begin{figure}
\centering
    \includegraphics[width=0.45\textwidth,keepaspectratio]{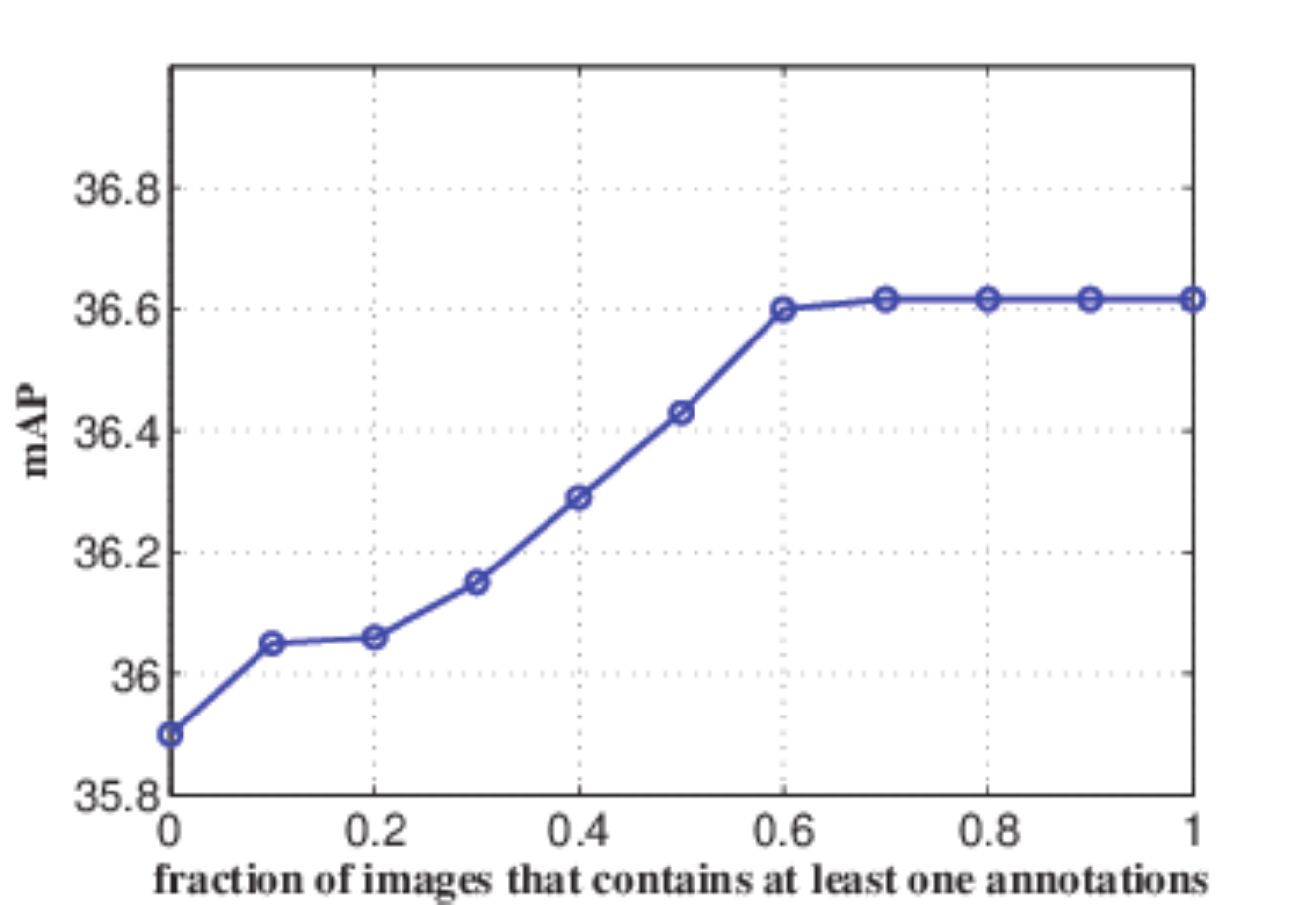}
  \caption{Adaptive distillation applied to ImageNet. Varying the fraction \(\rho\) of ImageNet\(_p\)-110.}\label{Imagent}
\end{figure}

\textbf{Results} We conducted the experiments using a combination of ImageNet\(_p\)-110 and ImageNet\(_n\)-110, as shown in Figure \ref{Imagent}. We randomly sample \(\rho\) fraction from ImageNet\(_p\)-110 and \((1-\rho)\) from ImageNet\(_n\)-110 so that the total number of images is still 110k. The more images sampled from ImageNet\(_p\)-110, the better performance the student will be, with \(\rho\) varing from 0 to 0.6. The proposed unlabeled data selection will improve the distillation performance when the fraction \(\rho\) is small. For the real-world data, the fraction is typically between 0 to 0.2.

\section{Conclusion}
In this paper, we design an adaptive distillation loss for the single-stage detector and demonstrate its effectiveness with RetinaNet in the common distillation setting . We also investigate this adaptive distillation in a semi-supervised learning setting. It is proved the student model can gain much improvement using both hard targets and soft targets produced by the teacher on unlabeled data. The student even surpass the teacher given enough transferring set. In the end, we demonstrate the proposed unlabeled data selection method is effective to transfer knowledge through unlabeled data with a dissimilar distribution compared with labeled data.

{\small
\bibliographystyle{ieee}

}

\end{document}